\documentclass[10pt,twocolumn,letterpaper]{article}

\usepackage{iccv}
\usepackage{times}
\usepackage{epsfig}
\usepackage{graphicx}
\usepackage{amsmath}
\usepackage{amssymb}
\usepackage{booktabs}
\usepackage{multirow}

\usepackage{caption} 
\usepackage[breaklinks=true,bookmarks=false]{hyperref}
\usepackage[capitalize]{cleveref}
\crefname{section}{Sec.}{Secs.}
\Crefname{section}{Section}{Sections}
\Crefname{table}{Table}{Tables}
\crefname{table}{Tab.}{Tabs.}

\iccvfinalcopy 

\def\iccvPaperID{1041} 

\begin{document}

\title{Revisiting Kernel Temporal Segmentation as an Adaptive Tokenizer for Long-form Video Understanding}

\author{Mohamed Afham \quad Satya Narayan Shukla \quad Omid Poursaeed \quad Pengchuan Zhang \\
Ashish Shah \quad Sernam Lim \\
Meta AI
}

\maketitle


\begin{abstract}
While most modern video understanding models operate on short-range clips, real-world videos are often several minutes long with semantically-consistent segments of variable length. A common approach to process long videos 
is applying a short-form video model over uniformly sampled clips of fixed temporal length and aggregating the outputs. This approach neglects the underlying nature of long videos since fixed-length clips are often redundant or uninformative. In this paper, we aim to provide a generic and adaptive sampling approach for long-form videos in lieu of the de facto uniform sampling. Viewing videos as semantically-consistent segments, 
we formulate a task-agnostic, unsupervised and scalable approach based on Kernel Temporal Segmentation (KTS) for sampling and tokenizing long videos. We evaluate our method on long-form video understanding tasks such as video classification and temporal action localization, showing consistent gains over existing approaches and achieving the state-of-the-art performance on long-form video modeling.      
\end{abstract}

\section{Introduction}
\label{sec:intro}

Majority of video understanding models are devised to learn representations of short-form videos ranging from 5 to 10 seconds \cite{slowfast, nonlocal, karpathy, Ng_2015_CVPR, mvitv1, vivit, mvitv2, videoswin}. 
These models usually suffer from computation and memory bottlenecks when processing videos of longer length. A common approach to overcome this bottleneck is to uniformly divide long videos into fixed-length clips, process each clip separately and aggregate the results. This approach is highly redundant as nearby clips often convey similar information and short clips that overlap semantically meaningful segments are often uninformative.  

Several works \cite{meng2020ar, ocsampler, nsnet, frameexit, scsampler} have previously investigated adaptive sampling to learn video representations in an efficient manner. 
These methods often devise a learnable adaptive sampler 
to select more representative frames of the video based on the reward or penalty provided by the final prediction score. However, these methods are often limited to the classification task and are heavily dependent on the specific tasks and datasets on which they are trained and cannot easily transfer to unseen tasks or datasets. 

Most of these adaptive sampling approaches are not scalable to sampling large number of frames which is required for understanding long-form videos. In fact, all the recent approaches \cite{vis4mer, lvu} for long-form video understanding use the de facto uniform sampling for sampling fixed-length clips from long videos.

In this work we propose a task-agnostic, adaptive and unsupervised sampling approach for long videos. Motivated by the intuition that humans perceive videos as semantically-consistent segments of variable length, we decompose the video to semantically meaningful segments using Kernel Temporal Segmentation (KTS) \cite{kts}. KTS extracts features from sparsely sampled candidate frames, computes the matrix of frame-to-frame similarity, and outputs a set of optimal change points corresponding to the boundaries of temporal segments. We then sample frames from each segment uniformly which comprises the input to the video understanding model.
Our KTS-based input tokenization achieves the following desirable attributes: (a) it is agnostic to the downstream task, (b) it yields semantically-consistent segments without relying on training data, and (c) it is scalable to arbitrary number of segments and frames for a given long video. 

We validate generalizability of KTS-based adaptive sampling on multiple downstream tasks and benchmarks. We evaluate KTS-based sampling for video classification on Breakfast \cite{breakfast} and LVU \cite{lvu} benchmarks achieving state-of-the-art performance. We also report results for temporal action localization on ActivityNet \cite{activitynet}, showing effectiveness of KTS-based sampling over standard uniform sampling. Furthermore, we provide a comparison with existing adaptive frame sampling methods on ActivityNet video classification and show that our approach outperforms the baselines.  

The main contribution of our work can be summarized as follows:
\begin{itemize}
    \item We propose an adaptive, unsupervised, and task-agnostic frame sampling mechanism for long videos based on Kernel Temporal Segmentation (KTS), which overcomes deficiencies of common sampling approaches.      
    \item We extensively evaluate KTS-based adaptive sampling against existing sampling techniques on video classification and temporal action localization tasks, showing consistent improvements and achieving state-of-the-art performance on long-form video understanding. 
\end{itemize}

\section{Related Work}
\label{sec:related}

Most of the video understanding models are devised to learn the representations of short-form videos ranging from $5$ to $10$ seconds \cite{slowfast, nonlocal, karpathy, Ng_2015_CVPR, mvitv1, vivit, mvitv2, videoswin}. While these approaches use various architectures such as 2D CNNs \cite{karpathy, cnnlstm, Ng_2015_CVPR}, 3D CNNs \cite{kinetics, slowfast, nonlocal, tran2019video} and Vision Transformers \cite{vivit, mvitv1, mvitv2, videoswin}, they often share uniform sampling for input tokenization. These models
usually suffer from computation and memory bottlenecks
when processing videos of longer length.

Recent approaches for long-form video modeling can be broadly divided into two categories: a) building specialized models for learning from long-form videos,  and b) adaptive sampling approaches for selecting frames from long-form videos. We discuss the related works in both the areas below:

\subsection{Long-form Video Understanding} 

Several works have been introduced to study the capability of video models in modeling videos of longer length. 
A movie based question answering dataset was introduced by Tapaswi \etal \cite{movieqa} and Bain \etal \cite{condensed} introduced a text-to-video retrieval benchmark based on videos from movies. However, those line of works explore the video-language learning ability of the model hence not ideal for video-only evaluation. 

Recent works \cite{videograph, timeception, lvu, vis4mer} improve the ability of learning long-range dependencies in the temporal domain of the videos in video classification setting. ViS4mer \cite{vis4mer} introduces a state-space sequence layer to model the extracted the short term clip features in a long video.  
The object transformer model in \cite{lvu} aims to capture the long-range interactions between tracked objects. 
Wu \etal \cite{lvu} recently introduced a long-video benchmark (LVU) comprised of 7 classification tasks and 2 regression tasks based on movie videos, which has become a standard benchmark for long-form video understanding. Another line of work focuses on temporal action localization (TAL) \cite{actionformer, tadtr} task which requires modeling the long-range dependencies and evaluated on long-video datasets such as ActivityNet \cite{activitynet} and Thumos \cite{thumos}. 

While the proposed approaches in video classification and temporal action localization show promising performance in the modeling aspect, uniform sampling is employed as the default input sampling strategy and hence, these approach require a large number of frames for understanding the long-form videos. Instead, in this work, we deploy the KTS-based adaptive sampling as input tokenization 
for both video classification and temporal action localization to study effectiveness of it over the standard uniform sampling. 

\subsection{Adaptive sampling} 
Several adaptive sampling based strategies \cite{adaframe, scsampler, mgsampler, ocsampler, nsnet, DSN, frameexit} have been proposed to overcome the computation issues faced by the standard uniform sampling in video classification. SCSampler \cite{scsampler} used a light-weight network to predict the saliency score of short-clips sampled uniformally in the long video. AdaFrame \cite{adaframe} introduces an LSTM network augmented with a global memory, to learn how to adaptively select frames conditioned on inputs for efficient video recognition. FrameExit \cite{frameexit} investigates an early existing procedure by employing a cascade of gating modules to automatically determine the earliest point in processing where an inference is sufficiently reliable. OCSampler \cite{ocsampler} effectively samples few frames from a selected number of frame candidates to dynamically localize and attend to
the instance-specific condensed clip of each video. Zhi \etal  design an adaptive sampling strategy named MGSampler \cite{mgsampler} aiming to choose a more effective input with a fixed length in trimmed and short. 

While our work is closely related to both OCSampler and MGSampler, in contrast to them KTS-based sampling is task-agnostic and applicable to various long-video understanding tasks (\textit{e.g.,} video recognition, temporal action localization). 
Unlike the prior adaptive sampling approaches, e.g. OCSampler \cite{ocsampler} where sampler is first trained separately, KTS-based sampling does not require two-stage training. KTS-based adaptive sampling is scalable and can be used to sample large number of frames from long-range videos effectively unlike the prior works which focus on sampling a small number of frames and are not easily scalable. 
KTS based sampling is also unsupervised and can be performed independently to the downstream task as it is based on the change points of the video features. 

\section{Method}
\label{sec:method}

\begin{figure*}
    \centering
    \includegraphics[width=0.99 \linewidth]{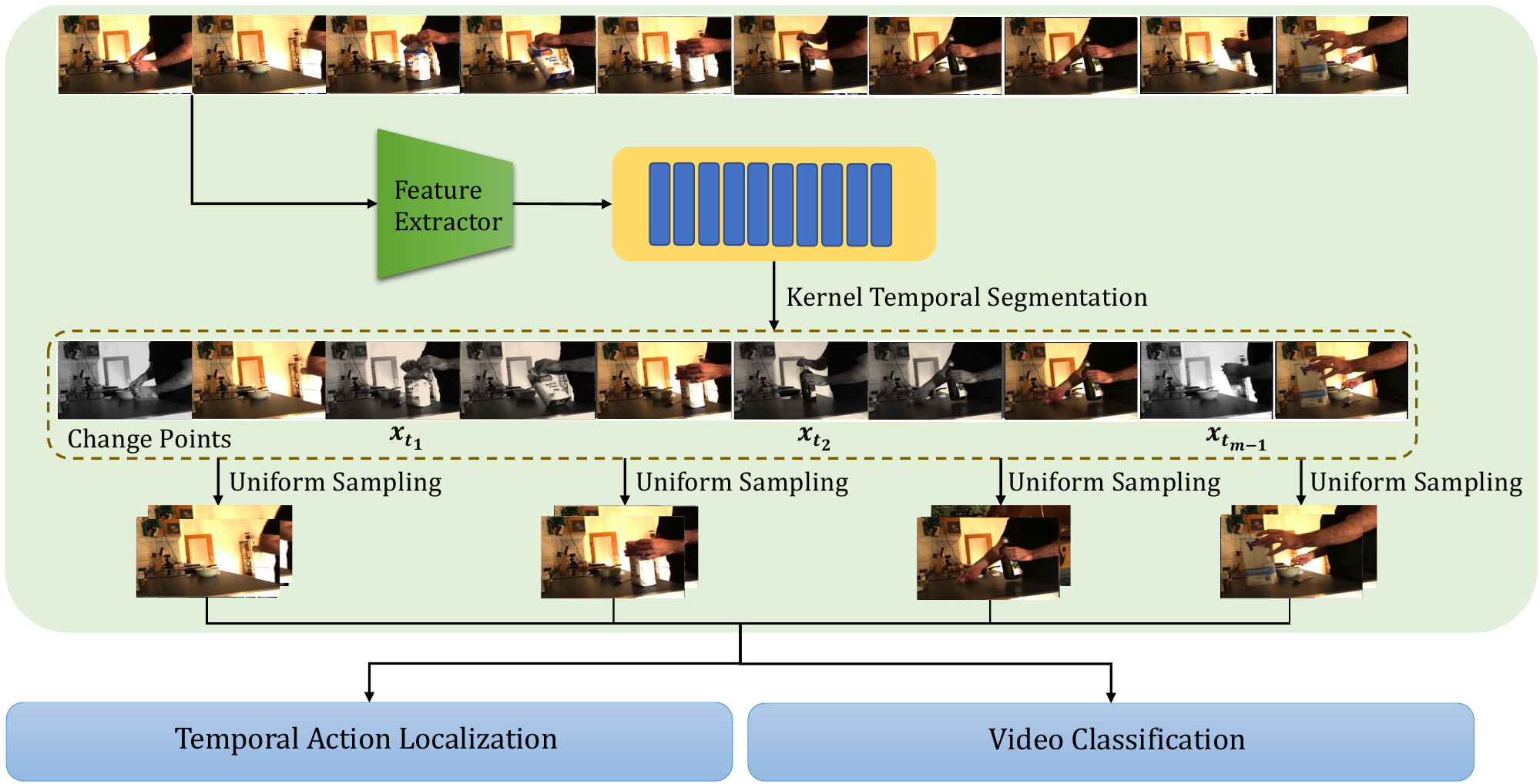}
    \caption{An overview of KTS-based adaptive sampling for Video Classification and Temporal Action Localization. The input video is initially downsampled and $m-1$ change points are computed using the KTS algorithm. $k$ frames are then uniformly sampled from each of the $m$ segments and are processed for the downstream task.}
    \label{fig:overall_architecture}
\end{figure*}

Conventional long-range video models process uniformly sampled short clips from the video and aggregate the results. However, relevant information in a long video is often not evenly distributed. Humans perceive videos as a sequence of coherent scenes/events and can have a semantic understanding of the scenes given a few sampled frames. Motivated by this intuition, we propose a similar approach for sampling and tokenizing long videos. We decompose videos into semantically consistent segments leveraging Kernel Temporal Segmentation (KTS) and sample frames uniformly from each segment. We first give an overview of the KTS algorithm in Sec. \ref{sed:kts}. Then we describe our sampling strategy for long-form video classification and action localization in Sec. \ref{sed:sampling}. Finally, we elaborate on the effectiveness of KTS over the other adaptive sampling techniques in the context of long-form video understanding.

\subsection{Kernel Temporal Segmentation}
\label{sed:kts}

The initial motivation behind KTS is to detect change points in the input and decompose the video into semantically consistent segments. 
KTS is a kernel-based algorithm that operates independently and in an unsupervised manner, hence it does not require any additional training to yield meaningful video segments. KTS has been extensively leveraged by several video summarization approaches \cite{mahasseni2017unsupervised, zhang2016video, rochan2018video, yao2016highlight, zhu2020dsnet} as the segmentation output provided by KTS has a significant impact on identifying highlights of the video and yielding a high-quality summarization of the video. Here we briefly describe the KTS algorithm. 

Given a long-form video, we initially downsample it, e.g. to one frame per second, and extract frame-level features using a pre-trained feature extractor $f_\theta$. Let $(x_i){^n_{i=1}}\in \mathbf{X}$ represent the sampled frames, $\mathbf{K}: \mathbf{X} \times \mathbf{X} \rightarrow{\mathbb{R}}$ represent a kernel function (Gram matrix) between descriptors $f_\theta(x_i)$ and $\phi: \mathbf{X} \rightarrow \mathcal{H}$ be the associated feature map with norm $\|.\|_\mathcal{H}$. 
Suppose we want to choose $m-1$ change points $x_{t_1}, \cdots, x_{t_{m-1}}$, which correspond to $m$ segments $[x_{t_0}, x_{t_1}], [x_{t_1}, x_{t_2}], \cdots, [x_{t_{m-1}}, x_{t_m}]$ with $x_{t_0}=0$ and $x_{t_m}=T$ being length of the video. 

The KTS algorithm minimizes the sum of the within-segment variances:
\begin{equation}
\label{eqn:var}
\min\limits_{m, t_1, \cdots, t_{m-1}}\sum^{m}_{i=1} var(t_{i-1}, t_{i})    
\end{equation}
where:
\begin{equation}
    var(t_{i-1}, t_i) = {\sum}_{t=t_{i-1}}^{t_{i}-1} \|\phi(x_t)-\mu_i\|^2
\end{equation}
and $\mu_i$ is the within-segment mean: 
\begin{equation}
\mu_i = \frac{{\sum}_{t=t_{i-1}}^{t_{i}-1} \phi{(x_t)}}{t_{i} - t_{i-1}}
\end{equation}
 
We can also make KTS adaptive to each video by making the number of segments $m$ variable. To avoid over-segmentation we add a penalty term $g(m, n)$ to the objective function. A common choice for $g(m, n)$ is $m\log(\frac{m}{n}+1)$. In this case, our final objective is: 
\begin{equation}
\label{eqn:var_penalty}
    \min\limits_{m, t_1, \cdots, t_{m-1}} \sum^{m}_{i=1} var(t_{i-1}, t_{i}) + g(m, n)
\end{equation}

In order to solve Equation \ref{eqn:var} and \ref{eqn:var_penalty}, we first compute the kernel for each pair of descriptors. We use a dot-product kernel in practice. Then the segment variances are computed for each possible starting point and segment duration. Finally, we use dynamic programming to minimize the objective and find the change points. Refer to \cite{kts} for more details. 

\subsection{Adaptive sampling with KTS}
\label{sed:sampling}

KTS algorithm yields a set of change points $x_{t_1}, \cdots, x_{t_{m-1}}$ which decompose the video into $m$ segments. Note that unlike shot boundary detection methods which focus on local differences between consecutive frames, KTS takes into account the differences between all pairs of frames. Therefore it provides semantically consistent and general segments. To represent each segment we uniformly sample $k$ frames from it. Long-form video models often consist of a backbone to process short-range clips and an aggregation mechanism (e.g. via a transformer or simple averaging).  
We feed sampled frames from each segment to the clip-level model which learns the representation for each segment/scene. The aggregation mechanism then combines scene-level information to obtain a global video-level representation. This is in line with how humans perceive videos. Despite its simplicity, we show that our sampling approach achieves state-of-the-art performance on long-form video modeling and outperforms existing samplers on several tasks and benchmarks. 

\subsection{Discussion}

As explained in Sec. \ref{sec:related} there are several other adaptive sampling techniques proposed in the literature. Our approach differs from these samplers in several ways. KTS-based sampling is generic and can be applied to various downstream tasks without training on them. However, existing samplers are task-specific and are often limited to video classification. We show in our experiments that our approach outperforms existing samplers on the video classification task. Unlike current approaches which disregard large portions of the video, KTS-based sampling minimizes loss of information as it samples from all the segments. This makes our approach well-suited for tasks such as action localization which need to preserve local information. 





\section{Experiments}
\label{sec:exp}
In this section, we present our experiments and results. We focus on video classification and temporal action localization tasks. 
We perform temporal action localization and classification on the ActivityNet \cite{activitynet} dataset, and video classification on the Breakfast dataset \cite{breakfast} and the LVU Benchmark \cite{lvu}. We also perform ablation experiments to show the impact of the number of frames used for video classification, number of change points estimated by KTS, and the backbone used as the feature extractor.

\subsection{Datasets}
{Breakfast} \cite{breakfast} is a human activity dataset focused on cooking-oriented actions. It comprises of $10$ categories of cooking breakfast. It contains $1712$ videos in total with $1357$ for training and $335$ for testing. The average length of a video is $2.3$ minutes. The cooking actions were performed by $52$ actors with $44$ for training and 8 for testing. This makes the task more challenging since the actors performing the actions during test time are not seen during training. \\

LVU (long-form video understanding benchmark) \cite{lvu} is compiled from the publicly available MovieClips dataset \cite{movie} which contains around 30,000 short movie snippets. The benchmarks consist of $9$ diverse tasks that require long-form video understanding. These tasks could be mainly categorized into \text{content understanding} \textit{(‘relationship', ‘speaking style', ‘scene/place')}, \text{movie metadata prediction} \textit{(‘director’, ‘genre’, ‘writer’, ‘movie release year’)} and \text{user engagement prediction} \textit{(‘YouTube like ratio’, ‘YouTube popularity’)}. Content understanding and movie metadata prediction can be considered as classification tasks hence are evaluated using the top-1 accuracy metric and user engagement prediction is a regression task and is evaluated using mean squared error (MSE). Each video is generally one to three minutes long. For both Breakfast and LVU datasets, we follow the training configuration suggested in ViS4mer. \\

ActivityNet \cite{activitynet} dataset contains around 20,000 untrimmed videos spanning 200 action classes of daily activities. The average length of a video is 117 seconds, and the average length of action segments is 48 seconds. Thus it can be considered as a long-form video dataset. Following the standard practice \cite{tsp, actionformer}, we train on the training split of the dataset and evaluate on the validation split. We report average $mAP@[0.5:0.05:0.95]$ similar to Actionformer \cite{actionformer} for fair comparison. 

\begin{table}[t]
\centering 
\caption{Video Classification results on Breakfast. We evaluate KTS-based sampling against uniform sampling with ViS4mer \cite{vis4mer} as the baseline. Our approach achieves state-of-the-art performance with significantly less computation.}
\label{table:breakfast}
\begin{tabular}{l c c c} 
\toprule 
Method & Frames & Accuracy\\
\toprule
VideoGraph \cite{videograph} & $64 \times 8$ & 69.50 \\
Timeception \cite{timeception} & $1024 \times 8$ & 71.30 \\
GHRM \cite{ghrm} & $64 \times 8$ & 75.49 \\
\hline
ViS4mer \cite{vis4mer} & $32 \times 32$ & $85.63$ \\
ViS4mer \cite{vis4mer} & $512 \times 32$ & $88.17$ \\
ViS4mer + KTS (Ours) & $\mathbf{32 \times 32}$ & {$\mathbf{89.86}$} \\
\bottomrule
\end{tabular}
\end{table}
\begin{table*}[t!]
\centering
\setlength{\tabcolsep}{8pt}
\caption{Evaluation of KTS-based sampling on the LVU benchmark. Our approach shows consistent improvements over uniform sampling on the majority of video understanding tasks.}
\label{tab:lvu}
\begin{tabular}{l|ccccccccc}
\bottomrule \multicolumn{1}{c|}{\multirow{2}{*}{Method}}  & 
\multicolumn{3}{c}{Content ($\uparrow$)} & \multicolumn{4}{c}{Metadata ($\uparrow$)} & \multicolumn{2}{c}{User ($\downarrow$)} \\ \cline{2-10}
& Relation & Speak & Scene & Director & Genre & Writer & Year & Like & Views\\ \toprule 
SlowFast + NL \cite{slowfast} & 52.40 & 35.80 & 54.70 & 44.90 & 53.00 & 36.30 & \textbf{52.50} & 0.38 & 3.77 \\
VideoBERT \cite{videobert} & 52.80 & 37.90 & 54.90 & 47.30 & 51.90 & 38.50 & 36.10 & 0.32 & 4.46 \\
Object Transformer \cite{lvu} & 53.10 & 39.40 & 56.90 & 51.20 & 54.60 & 34.50 & 39.10 & \textbf{0.23}& 3.55 \\
\hline
ViS4mer \cite{vis4mer} & 57.24 & 40.79 & 67.44 & 62.62 & 54.71 & 48.8 & 44.75 & 0.26 & 3.63 \\
ViS4mer + KTS (Ours) & \textbf{59.52} & \textbf{40.79} & \textbf{80.23} & \textbf{69.16} &  \textbf{65.86} & \textbf{54.16} & \underline{48.25} & 0.29 & \textbf{3.29} \\
\toprule
\end{tabular}
\end{table*}
\subsection{Video Classification}

We perform video classification experiments on Breakfast \cite{breakfast} and LVU \cite{lvu} datasets to study the effectiveness of KTS-based adaptive sampling.



\noindent \textbf{Baseline:} We adopt the recently introduced ViS4mer \cite{vis4mer} as the baseline model to evaluate the performance of KTS-based adaptive sampling against the uniform sampling on video classification tasks. ViS4mer is a long-range video classification model comprised of a standard Transformer encoder \cite{vit, videoswin} and a multi-scale temporal S4 \cite{s4} decoder. It extracts features from input video tokens using the Transformer encoder which are then fed to the multi-scale S4 decoder that learns hierarchical spatio-temporal video representations.
ViS4mer uses Vision Transformer \cite{vit} to extract features for experiments on the LVU benchmark and uses Video Swin Transformer \cite{videoswin} to extract features in experiments on the Breakfast dataset. Despite innovation in the modeling aspect, ViS4mer leverages uniform sampling to tokenize the input video. We adopt KTS-based adaptive sampling in both settings owing to its task-agnostic nature.


\noindent \textbf{Implementation Details:} Given a video, we downsample it to one frame per second and use the downsampled frames as candidates for computing the change points. We use GoogleNet \cite{googlenet} pre-trained on ImageNet-1K for extracting the feature descriptors. We sample $m \times k$ frames for each video as described in Sec. \ref{sed:sampling}, and the sampled frames are then fed to the video classification model.

\noindent \textbf{Results:} Table. \ref{table:breakfast} demonstrates the video classification results on the Breakfast dataset. We observe that KTS-based adaptive sampling achieves state-of-the-art results on the Breakfast dataset while utilizing $16\times$ fewer frames per video compared to the original ViS4mer baseline which uses uniform sampling. 
When compared with uniform sampling using the same setting $[32 \times 32]$, we observe a significant gain of $4.23 \%$ in terms of accuracy with KTS-based adaptive sampling, showing its superiority over uniform sampling. \\

Table. \ref{tab:lvu} shows the results on the LVU benchmark. KTS-based adaptive sampling achieves state-of-the-art performance on $7$ out of $9$ tasks and outperforms uniform sampling in $8$ out of $9$ tasks. In particular, in the \textit{scene prediction} under content understanding task of the benchmark, KTS-based tokenization yields a performance boost of $12.79\%$ while in the \textit{genre prediction} task under movie metadata prediction, KTS-based tokenization outperforms uniform sampling with a significant margin of $11.15\%$. Similar performance gains can be observed consistently throughout different tasks on the benchmark. 
Note that the LVU benchmark is more challenging than the Breakfast dataset. The tasks in LVU require long-term dependencies to be captured carefully, where the current short-range video models are proven to fail even with strong pre-training mechanisms \cite{lvu}. KTS-based input tokenization shows promising performance consistently throughout the benchmark which states the need to perform an adaptive sampling strategy to process long-form videos.

\subsection{Temporal Action Localization}
Temporal Action localization (TAL) aims to identify the action instances present in a video in the temporal domain and recognize the action categories. Despite the steady progress in TAL performance in the modeling aspects (\textit{e.g.,}  action proposals \cite{bmn}, pretraining \cite{tsp}, single-stage TAL \cite{actionformer}), uniform sampling is adopted as the de facto sampling approach in most of the action localization models. We analyze the impact of the KTS-based adaptive sampling mechanism on action localization.

\noindent \textbf{Baseline:} We investigate the performance of KTS-based sampling on the strong Actionformer \cite{actionformer} baseline, which achieves the current state-of-the art performance on TAL for ActivityNet. 
It comprises a multi-scale transformer encoder which encodes the sequence of embedded video clip features into a feature pyramid. The feature pyramid is then followed by a classification and a regression head to recognize the action instance and estimate the action boundaries respectively. TSP \cite{tsp} model pre-trained on ActivityNet video classification task is used to extract non-overlapping clip-level features. Refer to \cite{actionformer} for a complete description of Actionformer.

\begin{figure}[t]
    \centering
    \includegraphics[width = 0.9 \linewidth]{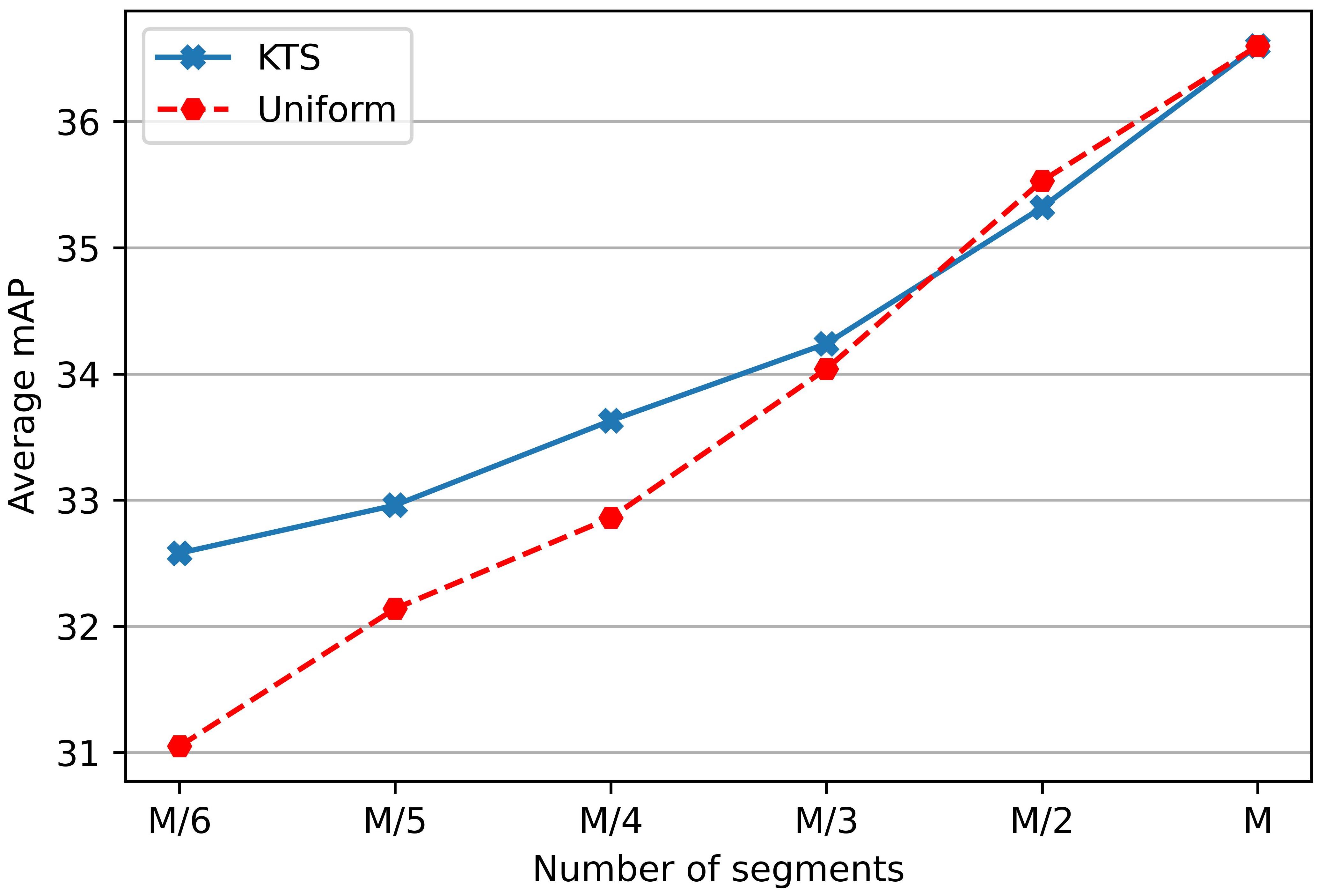}
    \caption{KTS vs Uniform sampling comparison on ActivityNet Action Localization. We report average mAP when varying the number of segments. $M$ corresponds to the number of segments when each segment length is $\frac{16}{15}$ seconds as used in the Actionformer baseline. }
    \label{fig:activitynet}
\end{figure}

\noindent \textbf{Implementation Details:} Given a video, we downsample it to one frame per second when computing the KTS change points and use ResNet-50 \cite{resnet} pre-trained on ImageNet-1K to extract feature descriptors for KTS computation. We adopt a similar training configuration as the Actionformer to study the impact of KTS-based adaptive sampling in TAL. Actionformer employs clips of 16 frames at a frame rate of $15$ fps and a stride of $16$ frames (i.e., non-overlapping clips) as input to the feature extractor followed by the localization module. This gives one feature vector per $\frac{16}{15} \approx 1.067$ seconds and $M = \frac{15}{16}T$ segments where $T$ is the video length. We can also consider $\frac{M}{2}$, $\frac{M}{4}$, $\cdots$ segments by sampling every $2^{nd}$, $4^{th}$, $\cdots$ frame. Similarly, we can choose $\frac{M}{2}$, $\frac{M}{4}$, $\cdots$ segments in our KTS-based sampling strategy. For the baseline, all the segments have the same length while our adaptive sampling technique yields variable-length segments. Within each segment, we uniformly sample $16$ frames in both cases. These frames are then fed to the action localization model. 
Fig. \ref{fig:activitynet} provides a comparison of KTS vs. uniform sampling, showing improved performance, especially for smaller number of segments.


\noindent \textbf{Results:} Fig. \ref{fig:activitynet} shows the empirical analysis of KTS-based sampling on TAL. Note that the performance gain of using KTS-based adaptive sampling is clearly observed for smaller number of segments (\textit{e.g.,} $\frac{M}{3}$ and below), and the gap in performance increases when reducing the number of segments. In particular, for $\frac{M}{6}$ segments uniform sampling achieves $31.05 \%$ average mAP while KTS-based sampling attains $32.58 \%$ average mAP on ActivityNet, yielding $1.53\%$ gain.  
For larger number of segments, the performance of KTS is nearly similar to uniform sampling. For $M$ segments, KTS reduces to uniform sampling as there are $M$ change point candidates when using one frame per second for sampling candidates. Similarly, for $\frac{M}{2}$ we select half of the candidates as change points, which makes it quite similar to uniform sampling. 


\subsection{Comparison with Existing Adaptive Sampling Methods}

\begin{table}[t]
    \scriptsize
    \centering
    \caption{Comparison of our approach with existing adaptive sampling strategies on ActivityNet video classification.}
    \label{tab:sampling}
    \begin{tabular}{lcccc}
        \toprule
        Method & Backbone  & mAP (\%) & GFLOPs \\ 
        \toprule
        NSNet \cite{nsnet} & ResNet-101 & 74.9 & 73.2 \\
        AdaFrame \cite{adaframe} & ResNet-101 & 71.5 & 78.7 \\
        LiteEval \cite{liteeval} & ResNet-101 & 72.7 & 95.1 \\
        KTS (Ours) $\left(84 \times 84\right)$ [8 frames] & ResNet-101 & \textbf{80.9} & \textbf{67.1}\\
        \midrule
        Uniform & ResNet-50 & 72.5 & 65.8 \\
        Random & ResNet-50 & 71.2 & 65.8 \\
        SCSampler \cite{scsampler} & ResNet-50 & 72.9 & 41.9 \\
        AdaMML \cite{panda2021adamml} & ResNet-50 & 73.9 & 94.0 \\
        AR-Net \cite{meng2020ar} & ResNet-50 & 73.8 & 33.5 \\
        ListenToLook \cite{listentolook} & ResNet-50 & 72.3 & 81.4 \\
        OCSampler \cite{ocsampler} & ResNet-50 & 79.8 & 67.2 \\
        KTS (Ours) $\left(84 \times 84\right)$ [6 frames] & ResNet-50 & 74.8  & \textbf{29.7} \\
        KTS (Ours) $\left(84 \times 84\right)$ [8 frames] & ResNet-50 & \textbf{80.0} & 32.1\\
        KTS (Ours) $\left(112 \times 112\right)$ [8 frames] & ResNet-50 & \textbf{80.3} & 37.4\\
        \bottomrule
    \end{tabular}
    \vspace{-5mm}
\end{table}

Table. \ref{tab:sampling} compares KTS-based adaptive tokenization with existing efficient frame sampling methods for video classification on the ActivityNet dataset. We use MobileNetv2 \cite{mobilenetv2} pre-trained on ImageNet-1K to extract the features. For a fair comparison with previous methods in terms of accuracy and computational cost, we initially uniformly sample 16 frames resized to a smaller resolution (e.g., 112 $\times$ 112) in a given video as the change point candidates and estimate change points. We sample one frame within each segment and train the ResNet50 classifier (pre-trained on Imagenet-1K) for video classification on ActivityNet. Our results show that KTS-based sampling yields a competitive performance when compared to existing adaptive sampling approaches. In particular, KTS-based sampling improves the classification accuracy by $1.03\%$ over AR-Net \cite{meng2020ar} while minimizing the computational cost by $3.8$ GFLOPS. KTS algorithm incurs only around $0.004$ GFLOPS in our experiments which is comparatively negligible to the computational cost incurred by ResNet50 and MobileNetV2. KTS-based sampling method also outperforms OCSampler \cite{ocsampler} while incurring significantly less computation cost.

 \begin{figure*}
    \centering
    \includegraphics[width = 0.78 \linewidth]{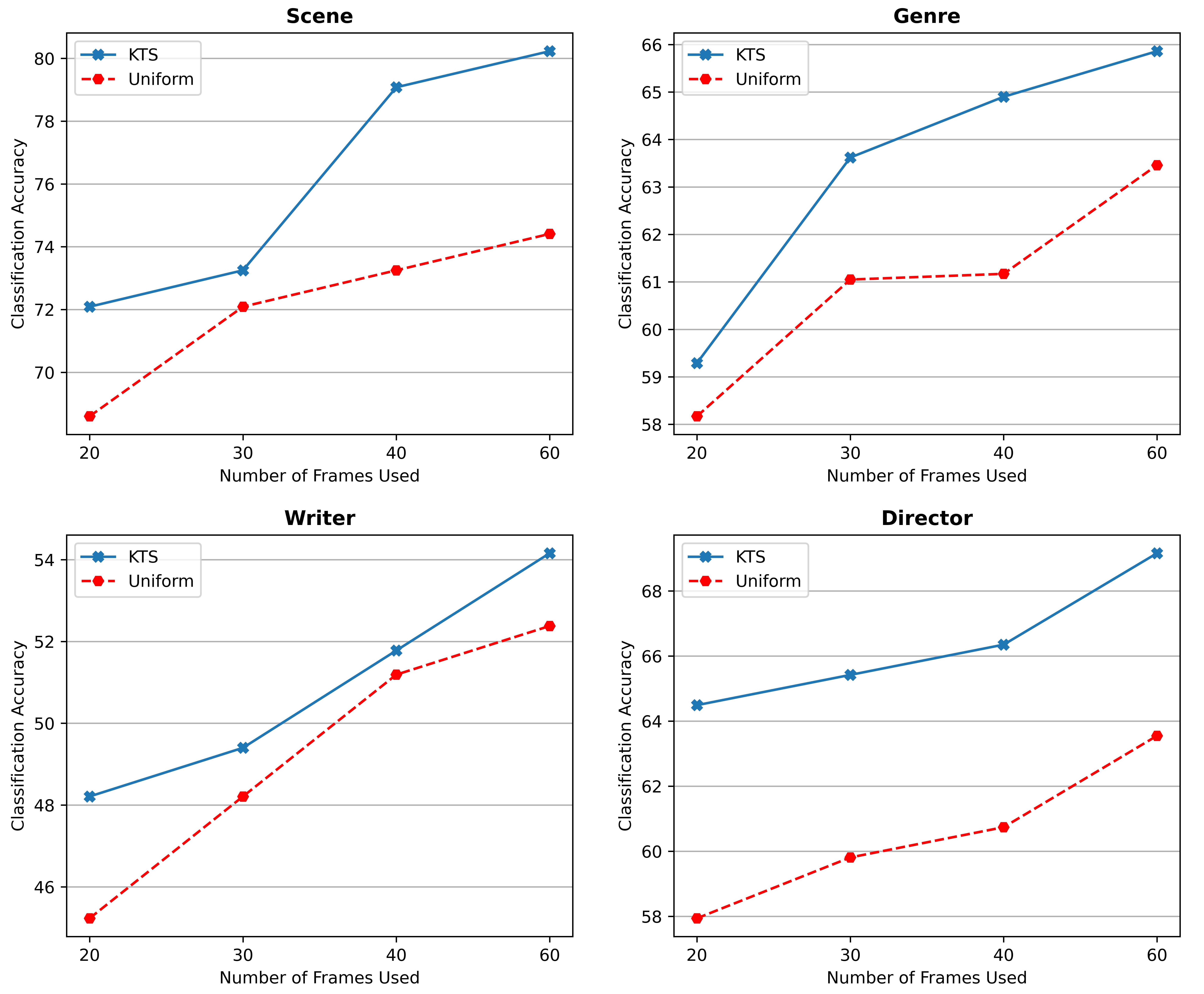}
    \caption{{KTS vs Uniform sampling comparison on classification tasks of the LVU benchmark by varying the number of input frames.} Consistent performance gain shows the effectiveness of KTS-based adaptive sampling over standard uniform sampling.}
    \label{fig:lvu}
\end{figure*}

\subsection{Ablation and Analysis}
In this section, we perform ablation experiments to show
the impact of the number of frames used for video classification, number of change points estimated by KTS, and the
backbone used as the feature extractor.

\begin{table}[t]
    \centering
    \caption{Impact of the number of change points in KTS-based adaptive video tokenization. We analyze the performance of our sampling approach on Breakfast video classification for different configurations of the number of change points. $m$: Number of video segments., $k$: Number of frames selected to process within each segment.}
    \label{tab:ncp}
    \begin{tabular}{clccc}
        \toprule
        Total Frames & $m \times k$ & Uniform & KTS & $\Delta$ \\ 
        \toprule
         \multirow{2}{*}{\textit{$256$}} & $16 \times 16$ & $78.87$ & $81.69$ & $+2.82\%$ \\
        & $32 \times 8$ & $81.13$ & $84.51$ & $+3.38\%$ \\
        \hline
        \multirow{3}{*}{\textit{$512$}} & $16 \times 32$ & $80.56$ & $84.51$ & $+3.95\%$ \\
        & $32 \times 16$ & $83.09$ & $88.17$ & $+5.08\%$\\
        & $64 \times 8$ & $83.66$ & $87.04$ & $+3.38\%$\\
        \hline
        \multirow{2}{*}{\textit{$1024$}} & $\mathbf{32 \times 32}$ & $\mathbf{85.63}$ & $\mathbf{89.86}$ & $\mathbf{+4.23\%}$  \\
        & $64 \times 16$ & $85.63$ & $86.76$ & $+1.13\%$ \\
        \bottomrule
    \end{tabular}
\end{table}



\begin{figure}[h]
    \centering
    \includegraphics[width = 0.95 \linewidth]{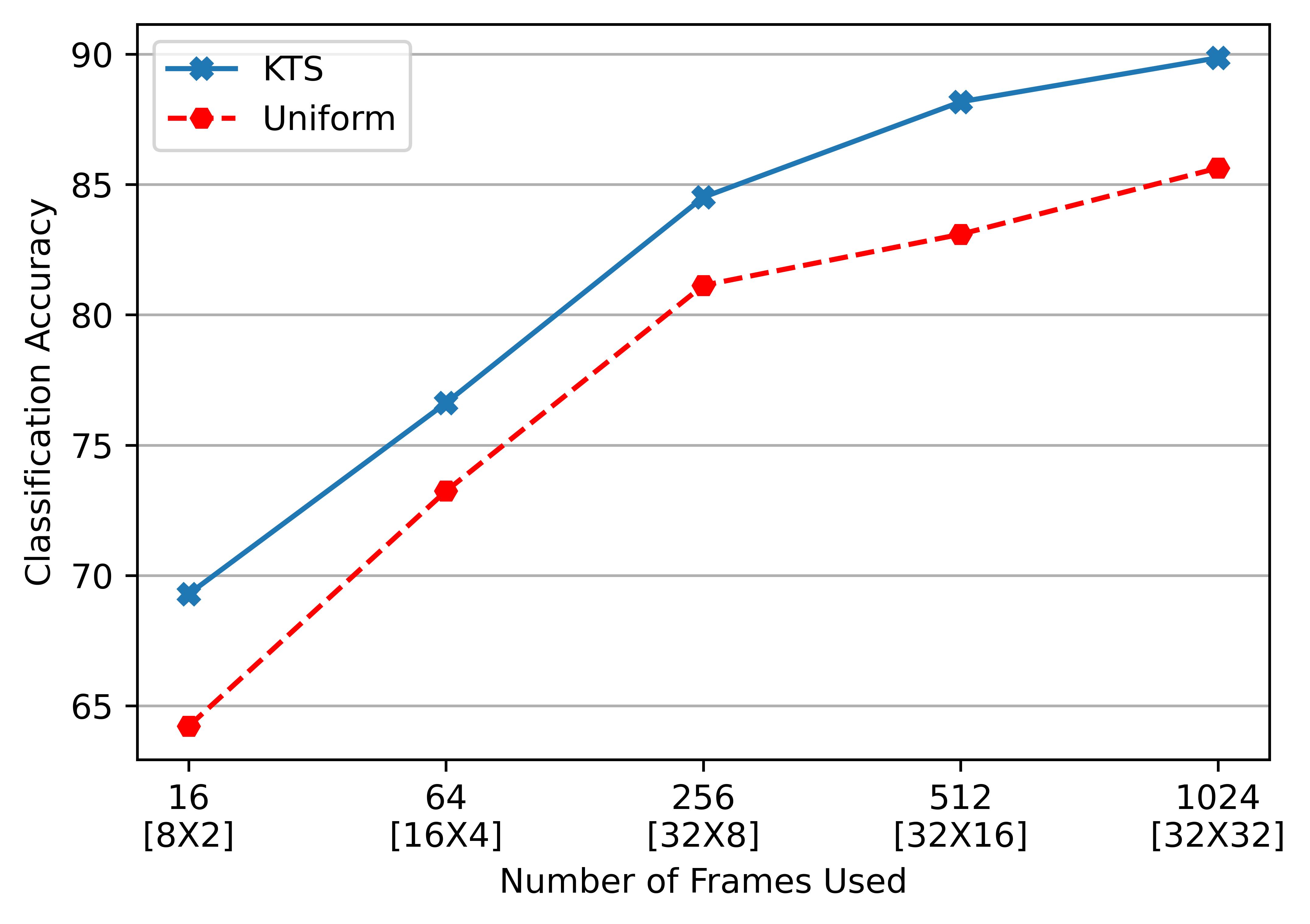}
    \caption{KTS vs Uniform sampling comparison on Breakfast video classification with a varying number of frames.} 
    \label{fig:breakfast}
\end{figure}

\begin{figure}[h]
    \centering
    \includegraphics[width = 0.9 \linewidth]{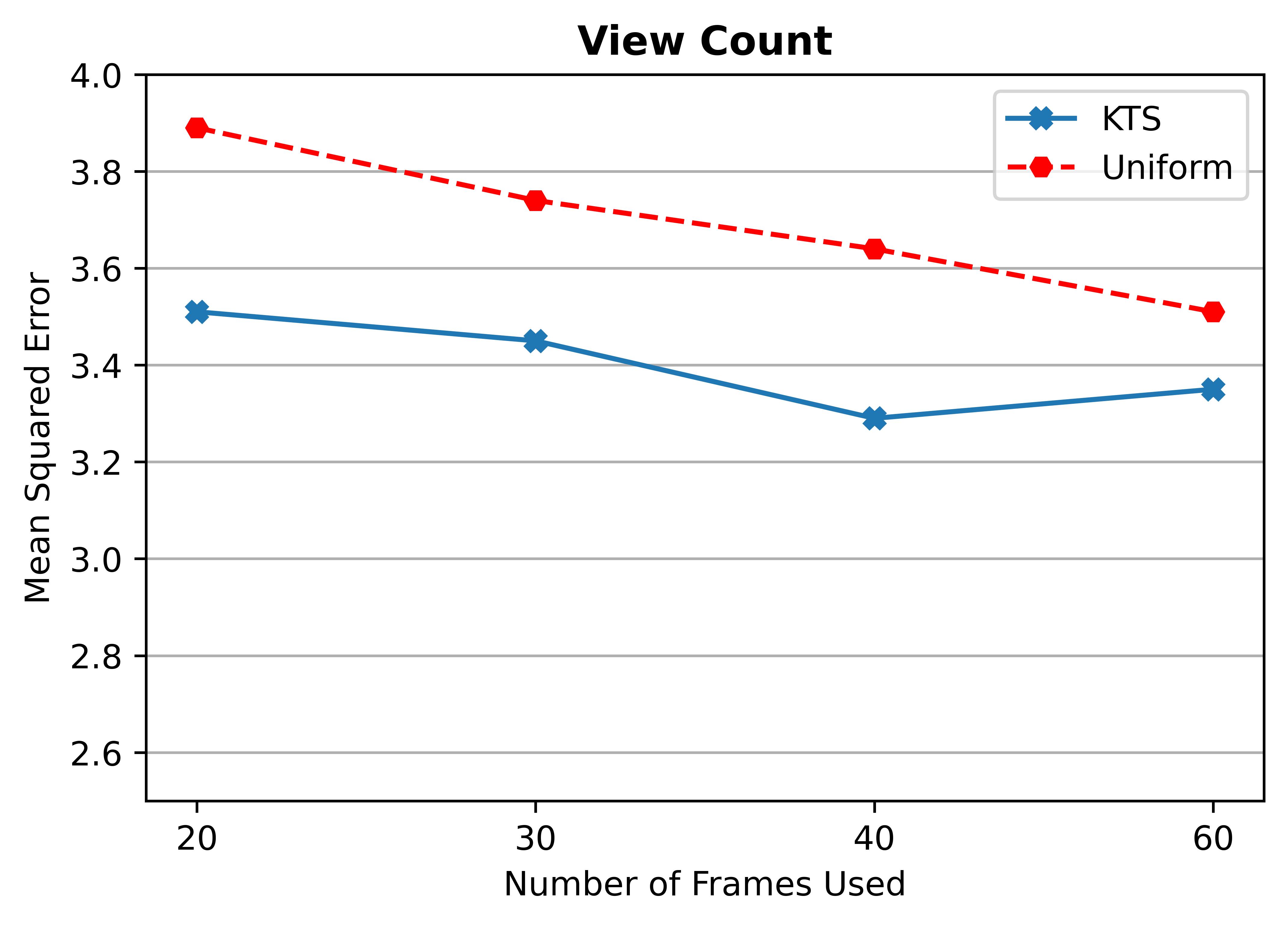}
    \caption{KTS vs Uniform sampling comparison on the view count prediction task of LVU benchmark.} 
    \label{fig:view}
\end{figure}

\subsubsection{Impact of the number of frames} 
We evaluate KTS-based adaptive sampling against uniform sampling on Breakfast and LVU dataset. We vary the number of input frames per video used for the classification using both the sampling approach. Fig. \ref{fig:lvu} presents the results on the LVU benchmark on four tasks: scene prediction, genre prediction, writer prediction, and director prediction. We vary the number of frames sampled for modeling. KTS-based adaptive sampling yields a consistent performance gain over standard uniform sampling in all configurations. 

Fig. \ref{fig:breakfast} demonstrates the results on Breakfast dataset.
KTS-based adaptive sampling shows consistent performance gain over uniform sampling in all settings. In particular, KTS-based sampling 
gives an accuracy of $84.51\%$ when using $256$ frames for each video in the $32 \times 8$ setting while uniformly sampling $512$ frames yields $83.09\%$ in the same setting. 
 This shows KTS-based tokenization not only improves the performance but also requires significantly less number of frames to process when compared to uniform sampling. 
 Table. \ref{table:breakfast} also validates this claim where KTS-based adaptive input tokenization achieves state-of-the art performance on Breakfast video classification with $(1/16)^{th}$ number of frames as that of uniform sampling. 
 
 Table. \ref{tab:ncp} reports the empirical results on different configurations of change points for a given number of frames input to the model. We observe that increasing the number of change points improves the performance of the model up to a certain point.   
 We also observe that KTS-based sampling consistently achieves significantly better performance compared to uniform sampling when compared in the same setting over all configurations. 
In particular, we obtain a performance gain of $4.23\%$ in the $32 \times 32$ setting where the number of change points estimated by the KTS algorithm is $31$. Similarly, $5.08\%$ boost is observed in $32 \times 16$ configuration over uniform sampling in the ViS4mer baseline.

Fig. \ref{fig:view} demonstrates the results on the regression task (view count prediction) of the LVU benchmark.  We vary the number of frames sampled for modeling. KTS-based adaptive sampling consistently achieves lower mean-squared error (MSE) compared to uniform sampling across different configuration settings. 

\subsubsection{Impact of the feature extractor on KTS} 
We investigate the choice of feature extractor in the KTS algorithm in Table. \ref{tab:feat}. 
We consider three standard image feature extractors namely: ResNet50 \cite{resnet}, MobileNetv2 \cite{mobilenetv2} and GoogleNet \cite{googlenet}. We use the pre-trained models after discarding the final classification layer. 
We observe that in both $32 \times 16$ and $64 \times 8$ settings, the GoogleNet backbone produces better results on the Breakfast dataset. 
Hence, similar to the choice of the number of frames, the choice of feature extractor plays a significant role in the KTS algorithm. 

\begin{table}[t]
    \centering
    \caption{Impact of the feature extractor of KTS on the Breakfast video classification task}
    \label{tab:feat}
    \begin{tabular}{lccc}
    \toprule
        $m \times k$ & ResNet50 & MobileNetv2 & GoogleNet \\ \toprule
        $64 \times 8$ & 81.12 & 83.38 & \textbf{87.04}\\
        $32 \times 16$ & 81.41 & 83.94 & \textbf{88.17} \\
        \bottomrule
    \end{tabular}
\end{table}

\section{Limitations and Future Work}
We identify two major limitations of KTS based input tokenization in long-form video understanding: 
(1) KTS, as a kernel based change point detection algorithm, is not learnable. The segmentation defects happening in KTS would directly affect the downstream task performance. However, learning change points specific to a task (\textit{e.g.,} video classification on Breakfast) can make transferability limited. We plan to investigate a learnable alternative to our current approach in the future. 
(2) While KTS performs superior to uniform sampling in several training configurations as shown in our analysis, the optimal choice of number of change points and the choice of feature extractor for KTS-based sampling still remains to be hand-picked. A possible future direction could be to investigate a learnable mechanism to select the optimal number of change points for a given video.

\section{Conclusion}

In this work, we present an adaptive and task-agnostic frame sampling mechanism for long-form video modeling. Our approach leverages Kernel Temporal Segmentation (KTS) to generate semantically-consistent segments used for sampling frames. We perform a comprehensive set of experiments on video classification and temporal action localization on several long-form video understanding datasets and benchmarks and show the superiority of KTS-based adaptive sampling against existing sampling strategies. In spite of its simplicity, our sampling approach achieves state-of-the-art performance on long-form video understanding benchmarks while being efficient. We plan to explore other variants of adaptive sampling based on temporal segmentation, which could operate in a learnable manner, in the future. 

{\small
\bibliographystyle{ieee_fullname}
\bibliography{egbib}
}

\end{document}


\title{Revisiting Kernel Temporal Segmentation as an Adaptive Tokenizer for Long-form Video Understanding - Supplementary Material}

\author{First Author\\
Institution1\\
Institution1 address\\
{\tt\small firstauthor@i1.org}
\and
Second Author\\
Institution2\\
First line of institution2 address\\
{\tt\small secondauthor@i2.org}
}

\maketitle
\ificcvfinal\thispagestyle{empty}\fi

\section*{Comparison with OCSampler \cite{ocsampler}}

In addition to the Table 3 in the main document, we compare the performance of KTS based sampling with OCSampler \cite{ocsampler} which shows the state-of-the-art performance among the existing adaptive sampling approaches. OCSampler aims to compress a long video to one short clip by sampling a selected number of frames from a given set of frame candidates. It introduces a network comprising of a skim network and a classifier network, where skim network is followed by a simple policy network which takes low resolution frame candidates as inputs and outputs a multinomial distribution of the relevancy of each candidate to the video representation and the classifier network processes the selected frames to make the final prediction. 

\begin{table}[h]
    \small
    \centering
    \caption{Comparison of our approach with OCSampler on ActivityNet video classification. Sampling Resolution denotes the resolution of frames used in feature extraction for frame sampling.}
    \label{tab:ocsampler}
    \begin{tabular}{lcccc}
        \toprule
        {\multirow{2}{*}{Method}}  & {\multirow{2}{*}{Frames}} & Sampling  & {\multirow{2}{*}{mAP (\%)}} & {\multirow{2}{*}{GFLOPs}} \\ 
        & & Resolution & & \\
        \toprule
        OCSampler \cite{ocsampler} & 16 & $112 \times 112$ & 79.82 & 67.2 \\
        KTS (Ours) & 16 & $112 \times 112$ & \textbf{80.33} & 73.6 \\
        KTS (Ours) & 16 & $84 \times 84$ & \textbf{79.93} & \textbf{66.8}\\
        \bottomrule
    \end{tabular}
\end{table}

Table.~\ref{tab:ocsampler} reports results of video classification on the ActivityNet dataset. We downsample the original frames to $112 \times 112$ and $84 \times 84$ resolution when estimating change points for KTS-based sampling. KTS-based sampling outperforms OCSampler on ActivityNet in terms of mean average precision (mAP) when 16 frames are sampled. Moreover, KTS-based sampling with $84 \times 84$ outperforms OCSampler on both mAP and computational cost (GFLOPs). \\

We also identify several attributes of KTS-based sampling where it overcomes the challenges faced by OCSampler.

\begin{itemize}
    \item Due to the complex reinforcement learning based sampling, OCSampler fails to scale beyond a certain range, and can only scale up to $\approx 16$ frames. However, KTS based sampling can scale for arbitrary number of frames per video.
    \item OCSampler employs a two-stage learning pipeline, where it initially warms up the skim network and the classifier network. Despite the efficient results, this two stage pipeline makes the model highly task-specific. However, KTS-based sampling is task-agnostic in nature, hence generalizable to several datasets and tasks.
\end{itemize}

{\small
\bibliographystyle{ieee_fullname}
\bibliography{egbib}
}